\documentclass[10pt,twocolumn,letterpaper]{article}

\usepackage[pagenumbers]{cvpr} 

\usepackage{graphicx}
\usepackage{amsmath}
\usepackage{amssymb}
\usepackage{booktabs}

\usepackage{booktabs}       
\usepackage{multirow}
\usepackage{pifont}
\usepackage[table]{xcolor}

\usepackage{array}
\newcolumntype{H}{>{\setbox0=\hbox\bgroup}c<{\egroup}@{}}
\newcommand{\cmark}{\ding{51}}%
\newcommand{\xmark}{\ding{55}}%
\makeatletter\renewcommand\paragraph{\@startsection{paragraph}{4}{\z@}
  {.2em \@plus1ex \@minus.2ex}{-.5em}{\normalfont\normalsize\bfseries}}\makeatother

\newcommand{\ourproposedmethod}{DeepFusion}
\newcommand{\lidar}{lidar}
\newcommand{\Lidar}{Lidar}
\newcommand{\camera}{camera}
\newcommand{\Camera}{Camera}
\newcommand{\InverseAug}{InverseAug}
\newcommand{\LearnableAlign}{LearnableAlign}

\definecolor{Highlight}{HTML}{39b54a}
\newcommand{\improves}[1]{{ \color{Highlight} {(#1)}}}
\newcommand{\Improves}[1]{{ \color{Highlight} (\textbf{#1})}}

\makeatletter
\newcommand{\ssymbol}[1]{^{\@fnsymbol{#1}}}
\def\blfootnote{\xdef\@thefnmark{}\@footnotetext}
\makeatother

\usepackage[pagebackref,breaklinks,colorlinks]{hyperref}

\usepackage[capitalize]{cleveref}
\crefname{section}{Sec.}{Secs.}
\Crefname{section}{Section}{Sections}
\Crefname{table}{Table}{Tables}
\crefname{table}{Tab.}{Tabs.}

\def\cvprPaperID{1664} 
\def\confName{CVPR}
\def\confYear{2022}

\begin{document}

\title{DeepFusion: Lidar-Camera Deep Fusion for Multi-Modal 3D Object Detection}

\author{
Yingwei Li\textsuperscript{\rm 1,2$\ast$}~~~~
Adams Wei Yu\textsuperscript{\rm 2$\ast$}~~~~
Tianjian Meng\textsuperscript{\rm 2}~~~~
Ben Caine\textsuperscript{\rm 2}~~~~
Jiquan Ngiam\textsuperscript{\rm 2}~~~~
Daiyi Peng\textsuperscript{\rm 2}~~~~ \\
Junyang Shen\textsuperscript{\rm 2}~~~~
Bo Wu\textsuperscript{\rm 2}~~~~
Yifeng Lu\textsuperscript{\rm 2}~~~~
Denny Zhou\textsuperscript{\rm 2}~~~~ 
Quoc V. Le\textsuperscript{\rm 2}~~~~
Alan Yuille\textsuperscript{\rm 1}~~~~
Mingxing Tan\textsuperscript{\rm 2} \\
\textsuperscript{\rm 1}Johns Hopkins University \qquad\qquad
\textsuperscript{\rm 2}Google \qquad\qquad
\\
 \{ywli, adamsyuwei, tanmingxing\}@google.com
\vspace{-.5em}

}

\maketitle

\begin{abstract}
\Lidar{}s and \camera{}s are critical sensors that provide complementary information for 3D detection in autonomous driving. While prevalent multi-modal methods~\cite{vora2020pointpainting,wang2021pointaugmenting} simply decorate raw \lidar{} point clouds with \camera{} features and feed them directly to existing 3D detection models, our study shows that fusing \camera{} features with deep \lidar{} features instead of raw points, can lead to better performance. However, as those features are often augmented and aggregated, a key challenge in fusion is how to effectively align the transformed features from two modalities. In this paper, we propose two novel techniques: \textbf{\InverseAug{}} that inverses geometric-related augmentations, \eg, rotation, to enable accurate geometric alignment between \lidar{} points and image pixels, and \textbf{\LearnableAlign{}} that leverages cross-attention to dynamically capture the correlations between image and \lidar{} features during fusion. Based on \InverseAug{} and \LearnableAlign{}, we develop a family of generic multi-modal 3D detection models named \textbf{\ourproposedmethod{}}, which is more accurate than previous methods. For example, \ourproposedmethod{} improves PointPillars, CenterPoint, and 3D-MAN baselines on Pedestrian detection for 6.7, 8.9, and 6.2 LEVEL\_2 APH, respectively. Notably, our models achieve state-of-the-art performance on Waymo Open Dataset, and show strong model robustness against input corruptions and out-of-distribution data. Code will be publicly available at \url{https://github.com/tensorflow/lingvo/tree/master/lingvo}. 
\blfootnote{$^\ast$Equal contribution. Work done when YL was an intern at Google.}
\end{abstract}


\begin{figure}[t]
\centering
\includegraphics[width=0.95\linewidth]{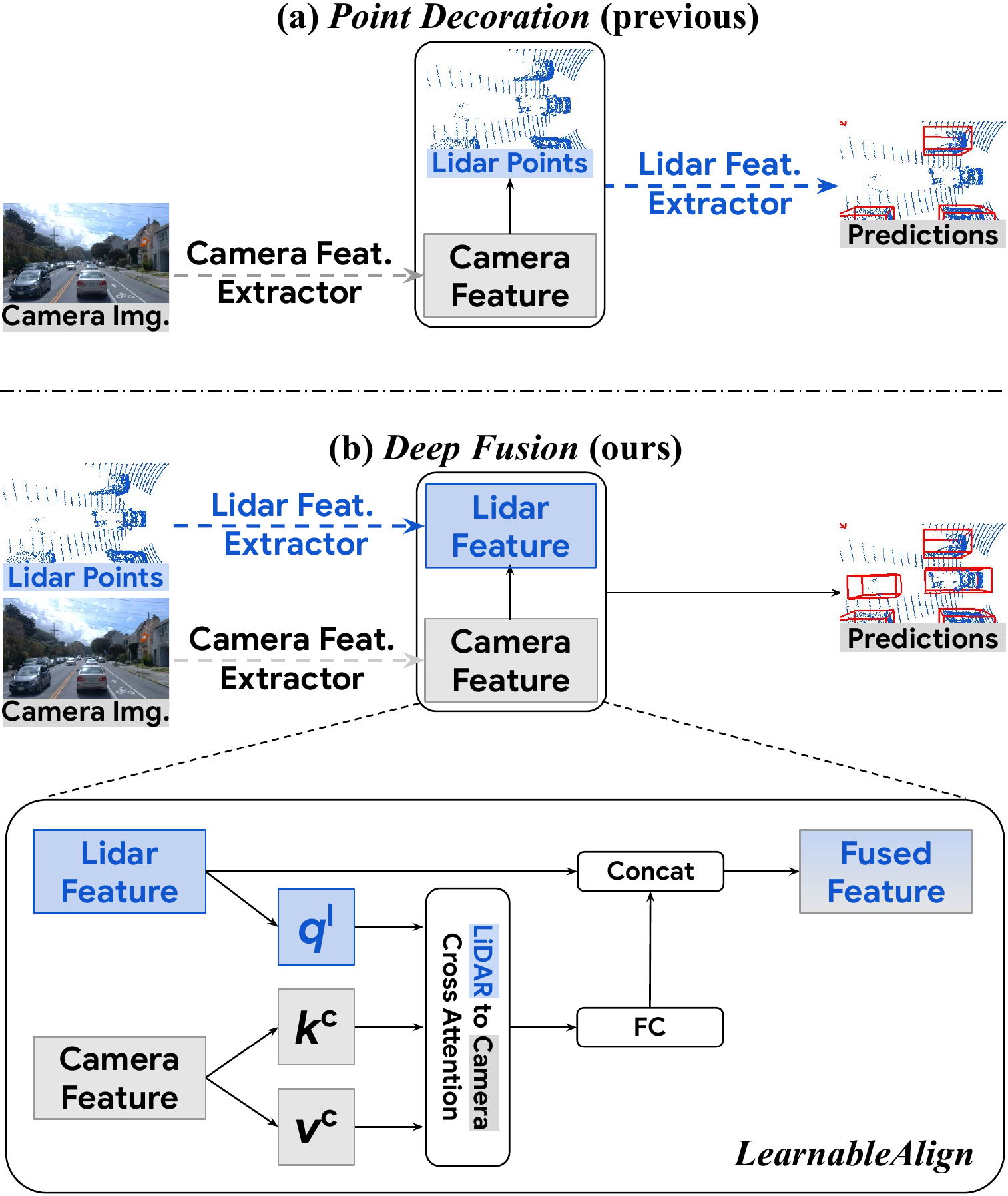}
\caption{Our method fuses two modalities on \textbf{deep feature level}, while previous state-of-the-art methods (PointPainting~\cite{vora2020pointpainting} and PointAugmenting~\cite{wang2021pointaugmenting} as examples) decorate \lidar{} points with \camera{} features on \textbf{input level}. To address the modality alignment issue (see Section~\ref{sec:intro}) for deep feature fusion, we propose two techniques \textbf{\InverseAug{}} (see Figure~\ref{fig:inverseaug} and~\ref{fig:inverseaug2}) and \textbf{\LearnableAlign{}}, a cross-attention-based feature-level alignment technique. } 
\label{fig:teaser}
\vskip -0.1in
\end{figure}

\section{Introduction} \label{sec:intro}

\Lidar{}s and \camera{}s are two types of complementary sensors for autonomous driving. For 3D object detection, \lidar{}s provide low-resolution shape and depth information, while \camera{}s provide high-resolution shape and texture information.  While one would expect the combination of both sensors to provide the best 3D object detector, it turns out that
most state-of-the-art 3D object detectors use only \lidar{} as the input (\href{https://waymo.com/open/challenges/2020/3d-detection/}{Waymo Challenge Leaderboard}, accessed on Oct 14, 2021). This indicates that how to effectively fuse the signals from these two sensors still remain challenging. 
In this paper, we strive to provide a generic and effective solution to this problem.

Existing approaches in the literature for fusing \lidar{}s and \camera{}s broadly follow two approaches (Figure \ref{fig:teaser}): they either fuse the features at an early stage, such as by decorating points in the \lidar{} point cloud with the corresponding camera features \cite{vora2020pointpainting,wang2021pointaugmenting}, or they use a mid-level fusion where the features are combined after feature extraction \cite{huang20epnet,liang18fusion}. One of the biggest challenges in both kinds of approaches is to figure out the correspondence between the \lidar{} and \camera{} features.  To tackle this issue, we propose two methods: \textit{\InverseAug{}} and \textit{\LearnableAlign{}} to enable effective mid-level fusion.
\textbf{\InverseAug{}} inverses geometric-related data augmentations (\eg, RandomRotation~\cite{zhou2018voxelnet}) and then uses the original \camera{} and \lidar{} parameters to associate the two modalities. 
\textbf{\LearnableAlign{}} leverages cross-attention to dynamically learn the correlation between a \lidar{} feature and its corresponding \camera{} features.
These two proposed techniques are simple, generic, and efficient. Given a popular 3D point cloud detection framework, such as PointPillars~\cite{lang2019pointpillars} and CenterPoint~\cite{yin2021center}, \InverseAug{} and \LearnableAlign{} help the \camera{} images effectively align with \lidar{} point cloud with marginal computational cost (\ie,~only one cross-attention layer). When fusing the aligned multi-modal features, the camera signals, with much higher resolution, significantly improve the model's recognition and localization ability. These advantages are especially beneficial for the long-range object detection.

We develop a family of multi-modal 3D detection models named \textbf{\ourproposedmethod{}s}, which offer the advantage that they (1) can be trained end-to-end and (2) are generic building blocks compatible with many existing voxel-based 3D detection methods. \ourproposedmethod{} serves as a plug-in that can be easily applied to most of the voxel-based 3D detection methods, such as PointPillars~\cite{lang2019pointpillars} and CenterPoint~\cite{yin2021center}.

Our extensive experiments demonstrate that (1) effective deep feature alignment is the key for multi-modal 3D object detection, (2) by improving alignment quality with our proposed \InverseAug{} and \LearnableAlign{}, \ourproposedmethod{} significantly improves the detection accuracy, and (3) compared with its single-modal baseline, \ourproposedmethod{} is more robust against input corruptions and out-of-distribution data.

On the Waymo Open Dataset, \ourproposedmethod{} improves several prevalent 3D detection models such as PointPillars~\cite{lang2019pointpillars}, CenterPoints~\cite{yin2021center}, and 3D-MAN~\cite{yang20213d} by 6.7, 8.9, and 6.2 LEVEL\_2 APH, respectively.
We achieve state-of-the-art results on Waymo Open Dataset that \ourproposedmethod{} improves 7.4 Pedestrian LEVEL\_2 APH over PointAugmenting~\cite{wang2021pointaugmenting}, the previous best multi-modal method, on the validation set.
This result shows that our method is able to effectively combine the \lidar{} and \camera{} modalities, where the largest improvements come from the recognition and localization for \textit{long-range} objects.

Our contributions can be summarized as three folds:
\begin{itemize}
    \item To our best knowledge, we are the first to systematically study the influence of deep feature alignment for 3D multi-modality detectors;
    \item  We propose \InverseAug{} and \LearnableAlign{} to achieve  deep-feature-level alignment, leading to accurate and robust 3D object detector;
    \item  Our proposed models, \ourproposedmethod{}s, achieve state-of-the-art performance on Waymo Open Dataset.
\end{itemize}

\section{Related Work}

\paragraph{3D Object Detection on Point Clouds.}

\Lidar{} point clouds are often represented as unordered set, and many 3D object detection methods tend to directly deal with such raw unordered points. PointNet~\cite{qi2017pointnet} and PointNet++~\cite{qi2017pointnet++} are early seminal works that directly apply neural networks on point cloud. Following them, \cite{ngiam2019starnet,PointRCNN-Shi-19,yang2018ipod,qi2018frustum} also learn features with PointNet-like~\cite{qi2017pointnet} layers. \Lidar{} point clouds can be also represented as dense range images, where each pixel contains extra depth information.  \cite{meyer2019lasernet,bewley2021range} directly work on the range images to predict 3D bounding boxes.

Another set of 3D detection methods convert \lidar{} points to voxels or pillars, leading to two more commonly used 3D detection methods: voxel-based and pillar-based method. VoxelNet~\cite{zhou2018voxelnet} proposes a voxel-based approach, which discretizes the point-cloud into a 3D grid with each sub-space is called a voxel. A dense 3D convlutional network can then be applied to this grid to learn detection features. SECOND~\cite{SECOND-Yan-18} builds on VoxelNet and proposes using sparse 3D convolutions to increase efficiency. Since 3D voxels are often expensive to process, PointPillars~\cite{lang2019pointpillars} and PIXOR~\cite{PIXOR-Yang-18} further simplify 3D voxels to bird-eye-view 2D pillars, where all voxels with the same z-axis are collapsed to a single pillar. These 2D pillars can then be processed with existing 2D convolutional detection networks to produce the bird-eye-view bounding boxes. Since 2D pillars are usually easy and fast to process, many recent 3D detection methods \cite{PillarBase-Wang-20, vora2020pointpainting, yin2021center, yang20213d} are built upon PointPillars. In this paper, we also choose PointPillar as our baseline approach for dealing with \lidar{} point clouds.

\paragraph{\Lidar{}-\camera{} Fusion.} Instead of relying on \lidar{} point cloud, monocular detection approaches directly predict 3D boxes from 2D images~\cite{chen2016monocular,ku2019monocular, qian2020end}. A key challenge for these approaches is 2D images do not have depth information, so most monocular detectors need to implicitly or explicitly predict depth for each 2D image pixel, which is often another very difficult task. Recently, there is a trend to combine \lidar{} and \camera{} data to improve 3D detection. Some approaches~\cite{qi2018frustum,wang2019frustum} first detect objects in 2D images then use the information to further process the point cloud. Previous works~\cite{chen2017multi,ku2018joint} also use a two-stage framework to perform object-centric modality fusion. 
In contrast to these methods, our approach is easier to plug-in into most existing voxel-based 3D detection methods.

\paragraph{Point Decoration Fusion.} PointPainting~\cite{vora2020pointpainting} proposes to augment each \lidar{} point with the semantic scores of \camera{} images, which are extracted with a pre-trained semantic segmentation network. PointAugmenting~\cite{wang2021pointaugmenting} points out the limitation of semantic scores, and proposes to augment \lidar{} points with the deep features extracted from a 2D object detection network on top of \camera{} images. As shown in Figure~\ref{fig:teaser}~(a), those methods rely on a pretrained module (\eg, 2D detection or segmentation model) to extract the features from the camera images, which are used to decorate the raw point clouds and then fed into a \lidar{} feature voxelizer to construct the bird-eye view pseudo images.

\paragraph{Mid-level Fusion.} Deep Continuous Fusion~\cite{liang18fusion}, EPNet~\cite{huang20epnet} and 4D-Net~\cite{piergiovanni20214d} attempt to fuse the two modalities by sharing the information between 2D and 3D backbones. However, an important missing piece in those works is an effective alignment mechanism between \camera{} and \lidar{} features, which is confirmed in our experiments to be the key for building an effective end-to-end multi-modal 3D object detector.
Even knowing the importance of effective alignment, we point out it is challenging to do so for the following reasons. First, to achieve the best performance on existing benchmarks, \eg, Waymo Open Dataset, various data augmentation strategies are applied to \lidar{} points and \camera{} images before the fusion stage. For example, RandomRotation~\cite{zhou2018voxelnet} that rotates the 3D world along z-axis, is usually applied to \lidar{} points but not applicable to \camera{} images, making it difficult for the subsequent feature alignment. 
Second, since multiple \lidar{} points are aggregated into the same 3D cube, i.e., \textit{voxel}, in the scene,  one voxel corresponds to a number of \camera{} features, and these \camera{} features are not equally important for 3D detection.

\section{\ourproposedmethod{}}
In Section~\ref{sec:pipeline}, we first introduce our deep feature fusion pipeline. Then, we conduct a set of preliminary experiments to quantitatively illustrate the importance of alignment for deep feature fusion in Section~\ref{sec:alignment_quality}. Finally, we propose two techniques, \InverseAug{} and \LearnableAlign{}, to improve the alignment quality in Section~\ref{sec:boost_alignment_quality}.

\subsection{Deep Feature Fusion Pipeline} \label{sec:pipeline}
As shown in Figure~\ref{fig:teaser}~(a), previous methods, such as PointPainting~\cite{vora2020pointpainting} and PointAugmenting~\cite{wang2021pointaugmenting}, usually use an extra well-trained detection or segmentation model as \camera{} feature extractor. For example, PointPainting uses Deeplabv3+\footnote{https://github.com/NVIDIA/semantic-segmentation} to generate per-pixel segmentation labels as camera features~\cite{vora2020pointpainting}. Then, the raw \lidar{} points are decorated with the extracted \camera{} features. Finally, the \camera{}-feature-decorated \lidar{} points are fed into a 3D point cloud object detection framework.

The pipeline above is improvable due to following reasons. First, the \camera{} features are fed into several modules that are specially designed for processing point cloud data. For example, if PointPillars~\cite{lang2019pointpillars} is adopted as the 3D detection framework, the camera features need to be voxelized together with the raw point clouds to construct bird's eye view pseudo images. However, the voxelization module is not designed for processing camera information. Second, the \camera{} feature extractor is learned from other independent tasks (\ie, 2D detection or segmentation), which may lead to (1) domain gap, (2) annotation efforts, (3) additional computation cost, and more importantly, (4) sub-optimal extracted features because the features are heuristically chosen rather than learned in an end-to-end manner.

To tackle the above two issues, we propose a deep feature fusion pipeline. To address the first problem, we fuse deep \camera{} and \lidar{} features instead of decorating raw \lidar{} points at the input level so that the \camera{} signals do not go through the modules designed for point cloud. For the second problem, we use convolution layers to extract \camera{} features and train these convolution layers together with other components of the network in an end-to-end manner. To summarize, our proposed deep feature fusion pipeline is shown in Figure~\ref{fig:teaser}~(b): the \lidar{} point clouds are fed into an existing \lidar{} feature extractor (\eg, Pillar Feature Net from PointPillars~\cite{lang2019pointpillars}) to obtain \lidar{} feature (\eg, the pseudo image from PointPillars~\cite{lang2019pointpillars}); the \camera{} images are fed into a 2D image feature extractor (\eg, ResNet~\cite{he2016deep}) to obtain \camera{} feature; then, the \camera{} feature is fused to the \lidar{} feature; finally, the fused feature is processed by the remaining components of the selected \lidar{} detection framework (\eg, Backbone and Detection Head from Pointpillars~\cite{lang2019pointpillars}) to obtain the detection results.

In contrast to previous designs, our method enjoys two benefits: (1) high-resolution \camera{} features with rich contextual information do not need to be wrongly voxelized and then converted from perspective view to bird's eye view; (2) domain gap and annotation issues are alleviated, and better \camera{} features can be obtained due to the end-to-end training. However, the disadvantages are also obvious: compared with input-level decoration, aligning \camera{} features with \lidar{} signals becomes less straightforward on deep feature level. For example, the inaccurate alignment caused by heterogeneous data augmentation for two modalities could pose a potential challenge for the fusion stage. In Section~\ref{sec:alignment_quality}, we verify that the misalignment can indeed harm the detection model, and provide our solution in Section~\ref{sec:boost_alignment_quality}.

\begin{table}[t]
    \centering
    \begin{tabular}{l|c|c|c|c}
    \hline
         Max Rotation & 0$^{\circ}$ & 15$^{\circ}$ & 30$^{\circ}$ & 45$^{\circ}$  \\
         \hline
         Single-Modal & 72.6 & 75.2 & 76.6 & 77.6 \\
         Multi-Modal & 75.2 & 76.1 & 77.3 & 78.0  \\
         Improvement & +2.6 & +0.9 & +0.8 & +0.4 \\
         \hline
    \end{tabular}
    \caption{Performance gain by multi-modal fusion diminishes as the magnitude of RandomRotation~\cite{zhou2018voxelnet} goes up, indicating the importance of accurate alignment. \InverseAug{} is not used here. On the Waymo Open Dataset pedestrian detection task, the LEVEL 1 AP improvements from single-modal to multi-modal are reported. 
    See Section~\ref{sec:alignment_quality} for more details.}
    \label{tab:alignment}
\end{table}

\subsection{Impact of Alignment Quality} \label{sec:alignment_quality}
To quantitatively assess the influence of alignment to deep feature fusion, we disable all other data augmentations but only twist the magnitudes of RandomRotation~\cite{zhou2018voxelnet} to the \lidar{} point cloud of our deep fusion pipeline during training. More details of the experimental settings can be found in the supplementary materials. Since we only augment the \lidar{} point cloud but keep the \camera{} images unchanged, stronger geometry-related data augmentation leads to worse alignment. As shown in Table~\ref{tab:alignment}, the benefit from multi-modal fusion diminishes as the rotation angle enlarges. For example, when no augmentation is applied (maximum rotation  $=0^{\circ}$), the improvement is the most significant (+2.6 AP); when maximum rotation is $45^{\circ}$, only +0.4 AP gain is observed. Based on these observation we conclude that alignment is critical for deep feature fusion that if the alignment is not accurate, the benefit from the camera input becomes marginal.

\begin{figure*}[t]
\centering
\includegraphics[width=1\linewidth]{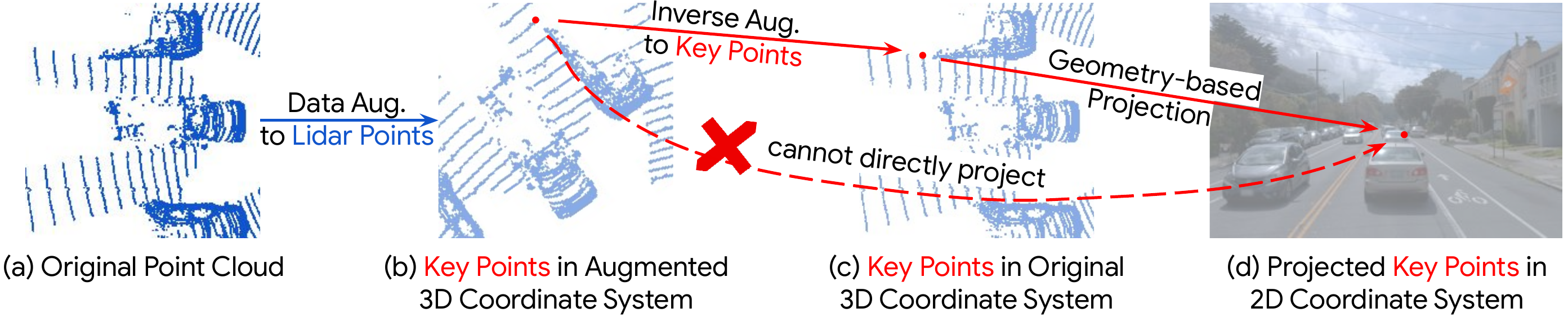}
\caption{\textbf{The pipeline of \InverseAug{}.} The goal of the proposed \InverseAug{} is to project the \textit{key points} obtained after the data augmentation stage,~\ie,~(a) $\rightarrow$ (b), to the 2D \camera{} coordinate system. The key point is a generic concept that can be any 3D coordinate, such as a \lidar{} point or a voxel center. For simplicity, we use a \lidar{} point here to illustrate the idea. It is less accurate to directly project the key points from the augmented 3D coordinate system to 2D camera coordinate system by using \camera{} and \lidar{} parameters,~\ie,~directly from (b) to (d). Here we propose to first find all key points in the original coordinate by inversely applying all data augmentation to the 3D key points,~\ie,~(b) $\rightarrow$ (c). Then, the \lidar{} and \camera{} parameters can be used to project 3D key points to \camera{} features,~\ie,~(c) $\rightarrow$ (d). \InverseAug{} significantly improves the alignment quality as shown in Figure~\ref{fig:inverseaug2}.}
\label{fig:inverseaug}
\end{figure*}

\begin{figure}[t]
\centering
\includegraphics[width=0.98\linewidth]{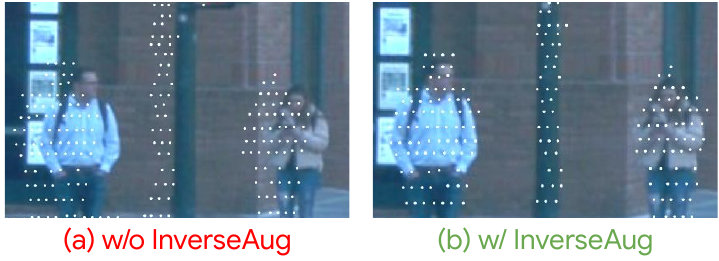}
\caption{The \camera{}-\lidar{} alignment quality comparison before and after applying \InverseAug{}. As shown in (a), without \InverseAug{}, the \lidar{} points (marked as white) are not well-aligned to the pedestrians and the pillar in \camera{} view. In contrast, as shown in (b), the \lidar{} points align better to the \camera{} data with \InverseAug{}. Note that we only add a small magnitude of data augmentation in this figure. The misalignment is more severe without \InverseAug{} during training where strong data augmentations are widely applied.} 
\label{fig:inverseaug2}
\end{figure}

\subsection{Boosting Alignment Quality} \label{sec:boost_alignment_quality}
In view of the importance of aligning deep features, we propose two techniques, \InverseAug{} and \LearnableAlign{}, to effectively align deep features from two modalities.

\paragraph{\InverseAug{}.}
To achieve the very best performance on existing benchmarks, most of the methods requires strong data augmentation, as the training usually falls into an overfitting scenario. The importance of data augmentation can be seen from Table~\ref{tab:alignment}, where the accuracy can be boosted by up to 5.0 for single-modal model. Besides, Cheng~\etal~\cite{cheng2020improving} also suggest the importance of data augmentation for training 3D object detection models.
However, the necessity of data augmentation poses a non-trivial challenge in our \ourproposedmethod{} pipeline. Specifically, the data from the two modalities are usually augmented with different augmentation strategies (\eg, rotating along z-axis for 3D point clouds combined with random flipping for 2D images), making the alignment challenging.

To address the alignment issue caused by geometry-related data augmentation, we propose \InverseAug{}. As shown in Figure~\ref{fig:inverseaug}, after data augmentation is applied to a point cloud, given a 3D \textit{key point} (which can be any 3D coordinate, such as \Lidar{} point, voxel center, \etc) in the augmented space, the corresponding \camera{} feature cannot be located in the 2D space by simply using the original \lidar{} and \camera{} parameters. 
To make the localization feasible, \InverseAug{} first saves the augmentation parameters (\eg, the rotation degree for RandomRotate~\cite{zhou2018voxelnet}) when applying the geometry-related data augmentation. At the fusion stage, it  reverses all those data augmentation to get the original coordinate for the 3D key point (Figure~\ref{fig:inverseaug}~(c)), and then finds its corresponding 2D coordinates in the \camera{} space. Note that our method is generic as it can align different types of key points(\eg, the voxel centers), although we only adopt \lidar{} points in Figure~\ref{fig:inverseaug} for simplicity, and it can also handle situations where both modalities are augmented. In contrast, existing fusion methods like PointAugmenting~\cite{wang2021pointaugmenting} can only deal with the data before augmentation. Finally, we show an example of the alignment quality improvement by \InverseAug{} in Figure~\ref{fig:inverseaug2}~(b).

\paragraph{\LearnableAlign{}.} 
For input-level decoration methods such as PointPainting~\cite{vora2020pointpainting} and PointAugmenting~\cite{wang2021pointaugmenting}, given a 3D \lidar{} point, the only  corresponding \camera{} pixel can be exactly located as there is a one-to-one mapping. In contrast, when fusing deep features in our \ourproposedmethod{} pipeline, each \lidar{} feature represents a voxel containing a subset of points and hence its corresponding \camera{} pixels are in a polygon. So the alignment becomes a 
one-voxel-to-many-pixels problem. A naive approach is to average over all pixels corresponding to the given voxel. 
However, intuitively, and as supported by our visualized results, these pixels are not equally important because the information from the \lidar{} deep feature unequally aligns with every \camera{} pixel. For example, some pixels may contain critical information for detection, such as the target object to detect, while others may be less informative, consisting of background such as roads, plants, occluders,~\etc.

In order to better align the information from \lidar{} features with the most related \camera{} features, we introduce \LearnableAlign{}, that leverages cross-attention mechanism to dynamically capture the correlations between two modalities as shown in Figure~\ref{fig:teaser}. 
Specifically, the input contains a voxel cell, and all its corresponding $N$ \camera{} features. \LearnableAlign{} uses three fully-connected layers to respectively transform the voxel to the query $q^{\text{l}}$, and \camera{} features to the keys $k^{\text{c}}$ and values $v^{\text{c}}$. For each query (\ie, voxel cell), we conduct inner product between the query and the keys to obtain the attention affinity matrix that contains $1 \times N$ correlations between the voxel and all its corresponding $N$ \camera{} features. Normalized by a softmax operator, the attention affinity matrix is then used to weigh and aggregate the values $v^\text{c}$ that contains \camera{} information. The aggregated \camera{} information is then processed by a fully-connected layer, and finally concatenated with the original \lidar{} feature. The output is finally fed into any standard 3D detection framework, such as PointPillars or CenterPoint for model training.

\section{Experiments}

We evaluate \ourproposedmethod{} on Waymo Open Dataset~\cite{sun2020scalability}, a large scale 3D object detection dataset for self-driving cars. Waymo Open Dataset contains 798 training, 202 validation, and 150 testing sequences. Each sequences have about 200 frames with \lidar{} points, \camera{} images, and labeled 3D bounding boxes. We evaluate and compare models using the recommended metrics, Average Precision (AP) and Average Precision weighted by Heading (APH), and report the results on both LEVEL\_1 (L1) and LEVEL\_2 (L2) difficulty objects. We highlight LEVEL\_2 APH in tables since it is the main metric for ranking in the Waymo Challenge Leaderboard.

\subsection{Implementation Details}
\paragraph{3D detection models.} 
We reimplement three popular point cloud 3D object detection methods, PointPillars~\cite{lang2019pointpillars}, CenterPoint~\cite{yin2021center}, and 3D-MAN~\cite{yang20213d}, as baselines. Besides, we also find that their improved versions (denote as PointPillars++, CenterPoint++, and 3D-MAN++) can be better baselines, which  use 3 layers of multilayer perceptron with hidden size 256 to construct pseudo image from point cloud inputs, and change the non-linear activation function from ReLU~\cite{hahnloser2000digital,nair2010rectified} to SILU~\cite{elfwing2018sigmoid,ramachandran2017searching}.
By default, all experiments are conducted with the 3D-MAN++ pedestrian models. The final model submitted to the test server also combined with other techniques such as model ensemble (noted as ``Ens''), which are discussed in Section~\ref{sec:app_imp_detail}.

\paragraph{\LearnableAlign{}.} We use fully connected layers with 256 filters to embed the \lidar{} feature and its corresponding \camera{} features. In the \lidar{} to \camera{} cross-attention module, a dropout operator with 30\% drop rate is applied to the attention affinity matrix as a regularization during training. The MLP layer after the cross-attention module is a fully connected layer with 192 filters. Finally, the concatenated feature is processed by another fully connected layer to squeeze the number of channels. Different from standard implementation of attention module, our implementation is combined with dynamic voxelization~\cite{zhou2020end}. Therefore, we put the TensorFlow-style pseudo code in the supplementary material, which contains more details of the implementation of \LearnableAlign{}.

\paragraph{\InverseAug{}.} Inspired by PPBA~\cite{cheng2020improving}, we sequentially apply the following data augmentation strategies to \lidar{} point cloud during training: RandomRotation $\rightarrow$
WorldScaling $\rightarrow$ GlobalTranslateNoise $\rightarrow$ RandomFlip $\rightarrow$ FrustumDropout $\rightarrow$ RandomDropLaserPoints. More details about the augmentation operators can be found in ~\cite{cheng2020improving}. Different from PPBA~\cite{cheng2020improving} and other works, here we save all randomly generated parameters for all geometry-related data augmentation (\ie,~RandomRotation, WorldScaling, GlobalTranslateNoise, RandomFlip). During the fusion stage, we inversely apply all these augmentation with saved parameters to all 3D key points to find the coordinates in the original 3D coordinate system before applying data augmentation. 
Besides, we also need to reverse the order of augmentation operations (\ie, RandomFlip $\rightarrow$ GlobalTranslateNoise $\rightarrow$ WorldScaling $\rightarrow$ RandomRotation).

\definecolor{rowhighlight}{rgb}{0, 150, 255}
\begin{table}[t]
\small
\begin{center}
\begin{tabular}{l@{\ \ }c@{\ \ }c@{\ \ }c@{\ \ }c@{\ \ }}
  \toprule 
   Method Name & AP/L1 & APH/L1 &  AP/L2 & \textbf{APH/L2} \\
      \cmidrule(r){1-1}\cmidrule(r){2-5}
  \textbf{\ourproposedmethod{}-Ens (ours)}\dag  & \textbf{84.37} & \textbf{83.22} & \textbf{79.54} & \textbf{78.41} \\
  InceptioLidar & 83.80 & 82.46 & 79.15 & 77.84 \\
  AFDetV2-Ens~\cite{hu2021afdetv2} & 84.07 & 82.63 & 79.04 & 77.64 \\
  Octopus\_Noah & 83.10 & 81.67 & 78.65 & 77.27 \\
  HorizonLiDAR3D~\cite{ding20201st}\dag & 83.28 & 81.85 & 78.49 & 77.11 \\
  \rowcolor{rowhighlight!25} \textbf{\ourproposedmethod{} (ours)}\dag$\ssymbol{1}$ & \textbf{81.89} & \textbf{80.48} & \textbf{76.91} & \textbf{75.54} \\
  Cascade3D & 81.17 & 79.63 & 75.84 & 74.36 \\
  INT & 80.29 & 78.81 & 75.30 & 73.89 \\
  IUI & 80.00 & 78.60 & 74.94 & 73.60 \\
  XMU & 80.53 & 78.77 & 75.14 & 73.45 \\
  Octopus-det & 79.25 & 77.75 & 74.63 & 73.20 \\
  \rowcolor{rowhighlight!25} AFDetV2~\cite{hu2021afdetv2}$\ssymbol{1}$ & 79.77 & 78.21 & 74.60 & 73.12 \\
  \footnotesize LENOVO\_LR\_PCIE\_Det & 79.46 & 78.07 & 74.31 & 72.97 \\
  \footnotesize LENOVO\_LR\_PCIE\_RT\_Det & 79.42 & 77.98 & 74.31 & 72.97 \\
  \rowcolor{rowhighlight!25} CenterPoint++~\cite{yin2021center}$\ssymbol{1}$ & 79.41 & 77.96 & 74.22 & 72.82 \\
  \rowcolor{rowhighlight!25} SST\_v1~\cite{fan2021embracing}$\ssymbol{1}$ & 79.99 & 78.31 & 74.41 & 72.81 \\
  \bottomrule
 
 \end{tabular}
\end{center}
\vskip -0.15in
\caption{Leaderboard of the Waymo Open Dataset Challenge on 3D detection track. $\ssymbol{1}$: to our best knowledge, these entries (highlighted by \colorbox{rowhighlight!25}{light blue}) do not use model ensemble. \dag: multi-modal methods. }
\label{tab:waymo_test}
\end{table}

\begin{table}[t]
\setlength\tabcolsep{20pt}
\small
\begin{center}
\begin{tabular}{@{}l@{\ \ }l@{\ \ }c@{\ \ }c@{\ \ }c@{\ \ }c@{}}
  \toprule 
   Diff- & \multirow{2}{4em}{Method} & \multicolumn{2}{c}{Veh.} & \multicolumn{2}{c}{Ped.}  \\
   iculty & & AP & APH &  AP & APH \\
       \cmidrule(r){1-2}\cmidrule(r){3-4}\cmidrule(r){5-6}
       \multirow{11}{3em}{L1}  
  & PointPillars~\cite{lang2019pointpillars,sun2021rsn} & 63.3 & 62.7 & 68.9 & 56.6  \\  
  & PPBA~\cite{cheng2020improving,cheng2020improving} & 62.4 & \_ & 66.0 & \_  \\   
  & MVF~\cite{zhou2020end} & 62.9 & \_ & 65.3& \_  \\ 
  & CVCNet~\cite{chen2020every} & 65.2 & \_ & \_ & \_ \\ 
  & \small PointAugmenting~\cite{wang2021pointaugmenting}\dag & 67.4 & \_ & 75.4 & \_ \\
  & Pillar-OD~\cite{wang2020pillar} & 69.8 & \_ & 72.5& \_  \\ 
  & PV-RCNN~\cite{PVRCNN-Shi-20} & 74.4 & 73.8 & 61.4& 53.4  \\
  & CenterPoint~\cite{yin2021center} & 76.7 & 76.2 & 79.0 & 72.9  \\
  & RSN~\cite{sun2021rsn} & 78.4 & 78.1 & 79.4 & 76.2 \\
  & \ourproposedmethod{} (ours)\dag & \underline{80.6} & \underline{80.1} & \underline{85.8} & \underline{83.0} \\
  & \ourproposedmethod{}-Ens (ours)\dag & \textbf{83.6} & \textbf{83.2} & \textbf{87.1} & \textbf{84.7} \\
  \cmidrule(r){1-2}\cmidrule(r){3-4}\cmidrule(r){5-6}
  \multirow{6}{3em}{L2}  
  & PointPillars~\cite{lang2019pointpillars,sun2021rsn} & 55.2 & 54.7 & 60.0 & 49.1  \\  
  & \small PointAugmenting~\cite{wang2021pointaugmenting}\dag & 62.7 & \_ & 70.6 & \_ \\
  & PV-RCNN \cite{PVRCNN-Shi-20} & 65.4 & 64.8 & 53.9 & 46.7 \\ 
  & CenterPoint~\cite{yin2021center} &  68.8 & 68.3 & 71.0 & 65.3 \\ 
  & RSN~\cite{sun2021rsn} & 69.5 & 69.1 & 69.9 & 67.0 \\
  & \ourproposedmethod{} (ours)\dag  & \underline{72.9} & \underline{72.4} & \underline{78.7} & \underline{76.0} \\
  & \ourproposedmethod{}-Ens (ours)\dag & \textbf{76.0} & \textbf{75.6} & \textbf{80.4} & \textbf{78.1} \\
  \bottomrule
 
 \end{tabular}
\end{center}
\vskip -0.15in
\caption{Performance comparison between models for 3D detection on Waymo validation set. \dag: multi-modal methods. 
}
\label{tab:waymo_val}
\vskip -0.2in
\end{table}

\subsection{State-of-the-art performance on Waymo Data}
We compare our method with the published and unpublished 3D object detection methods on Waymo Open Dataset {\tt validation} and {\tt test} sets. 

According to the test results in Table~\ref{tab:waymo_test}, \ourproposedmethod{} achieves the best results on Waymo Challenge Leaderboard demonstrating the effectiveness of our approach. For example, \ourproposedmethod{}-Ens achieves the best results on the Waymo Challenge Leaderboard; \ourproposedmethod{} improves 2.42 APH/L2 compared with previous state-of-the-art single-model method, AFDetV2~\cite{hu2021afdetv2}.

We also compare different methods on the validation set as shown in Table~\ref{tab:waymo_val}. \ourproposedmethod{} significantly outperforms existing methods, demonstrating the effectiveness of our approach.

\subsection{\ourproposedmethod{} is a generic fusion method} Now we examine how generic our method is by plugging it to prevalent 3D detection frameworks.  
We conduct six pairs of comparison, each of which is between a single-modal method and its multi-modal counterpart. Those six \lidar{}-only models are PointPillars, CenterPoint, 3D-MAN and their improved version (marked as ``++"). As shown in Table~\ref{tab:validation_compare}, \ourproposedmethod{} shows consistent improvement for all single-modal detection baselines. 
These results indicate \ourproposedmethod{} is generic and can be potentially applied to other 3D object detection frameworks.

\begin{table}[!h]
\setlength\tabcolsep{5.2pt}
\small
\begin{center}
\begin{tabular}{l|c|cccc}
\hline
\multirow{2}{*}{Model}   & \multirow{2}{*}{Modal} & \multicolumn{2}{c}{LEVEL 1} & \multicolumn{2}{c}{LEVEL 2} \\
  &  & AP  & APH & AP & \textbf{APH} \\\hline
PointPillars\cite{lang2019pointpillars} & L & 67.4 & 55.6 & 58.4 & 48.1~~~~~~~~~~~ \\
\footnotesize \: +\ourproposedmethod{} & L + C & 72.0 & 62.8 & 63.0 & 54.8\improves{+6.7} \\\hline 
PointPillars++ & L & 72.5 & 62.9 & 63.3 & 54.8~~~~~~~~~~~ \\ 
\footnotesize \: +\ourproposedmethod{} & L + C & 73.9 & 65.1 & 64.9 & 57.0\improves{+2.2} \\\hline
CenterPoint\cite{yin2021center} & L & 71.9 & 62.0 & 63.1 & 54.3~~~~~~~~~~~ \\
\footnotesize \: +\ourproposedmethod{} & L + C & 78.3 & 70.8 & 70.2 & 63.2\improves{+8.9} \\\hline
CenterPoint++ & L & 77.5 & 69.5 & 68.5 & 61.3~~~~~~~~~~~ \\
\footnotesize \: +\ourproposedmethod{} & L + C & 81.2 & 75.0 & 73.1 & 67.2\improves{+5.9} \\\hline
3D-MAN~\cite{yang20213d} & L & 71.3 & 58.9 & 63.4 & 52.2~~~~~~~~~~~ \\
\footnotesize \: +\ourproposedmethod{} & L + C & 75.9 & 65.9 & 67.4 & 58.4\improves{+6.2} \\\hline
3D-MAN++ & L & 78.5 & 70.2 & 70.7 & 63.0~~~~~~~~~~~ \\
\footnotesize \: +\ourproposedmethod{} & L + C & 81.2 & 75.0 & 72.8 & 67.0\improves{+4.0} \\\hline
\end{tabular}
\end{center}
\vspace{-0.1in}
\caption{Plugging \ourproposedmethod{} into different single-modal baselines on Waymo validation set. \texttt{L} denotes \lidar{}-only; \texttt{L+C} denotes \lidar{} + \camera{}.
We evaluate Pointpillar, CenterPoint, 3D-MAN,  and their improved versions (denoted with ``++''). By adding \camera{} information, our \ourproposedmethod{} consistently improves the quality over \lidar{}-only models.}
\label{tab:validation_compare}
\end{table}

\subsection{Where does the improvement come from?}
\begin{figure}[tb]
\centering
\includegraphics[width=1\linewidth,trim=1.2cm 0.8cm 0.5cm 0.5cm,clip]{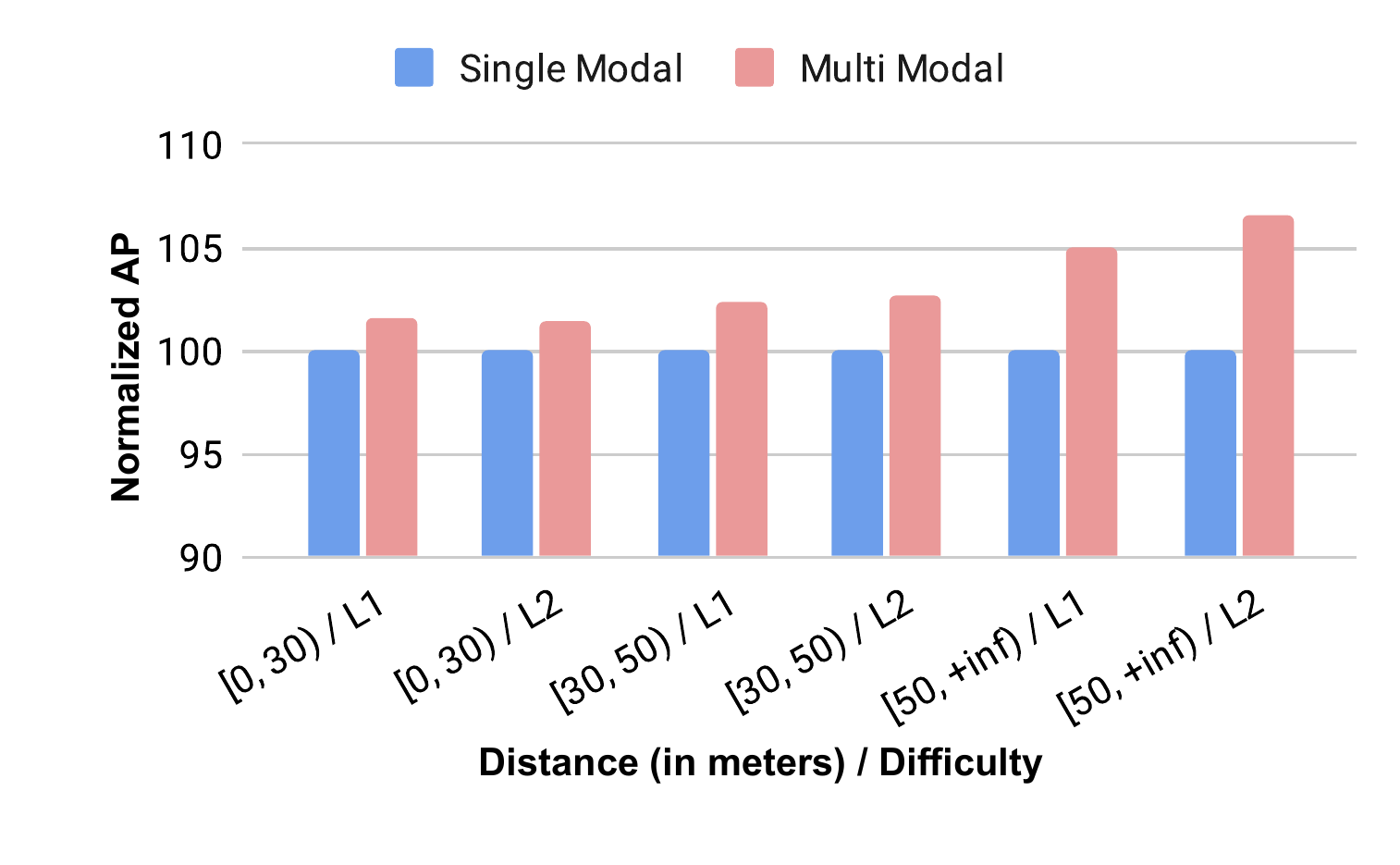}
\vskip -0.1in
\caption{Comparison between the single-modal baseline and \ourproposedmethod{} by showing the AP metric (with all the blue bars normalized to 100\%) across different ground-truth depth ranges. The results show \ourproposedmethod{} marginally improves the performance on short-range objects (\eg, within 30 meters) but significantly improves the performance on long-range objects (\eg, beyond 50 meters).} 
\label{fig:breakdown}
\end{figure}

\begin{figure*}[tb]
\centering
\includegraphics[width=0.9\linewidth]{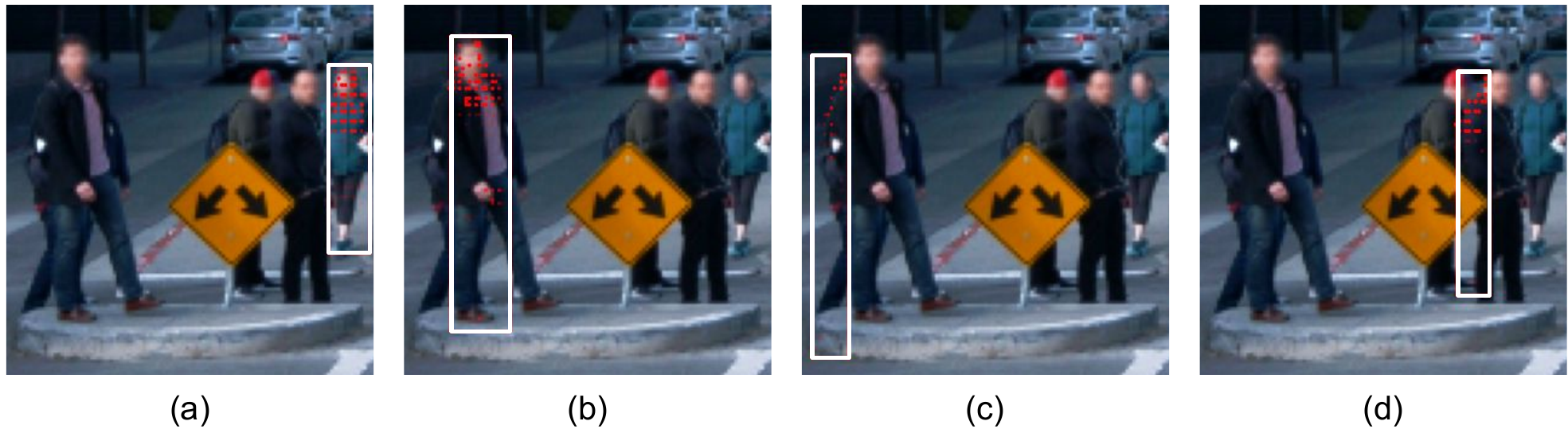}
\vskip -0.1in
\caption{The the attention map visualization for \LearnableAlign{}. For each subfigure, we study one 3D point pillar, which is marked by the white box in the 2D image. The important region indicated by the attention map is marked by red points. We have two interesting observations: first, as shown in (a) and (b), \LearnableAlign{} usually attends to the heads of the pedestrians, probably because the head is a discriminative part of a human from \camera{} image (it is difficult to recognize a head according to \lidar{} signals); second, as shown in (c) and (d), \LearnableAlign{} also  attends to object extremities (such as the back), which strives to use the high-resolution \camera{} information for predicting the object boundary to get accurate object size.}
\label{fig:vis_attention}
\vskip -0.1in
\end{figure*}

To better understand how \ourproposedmethod{} utilizes the \camera{} signal to improve the 3D object detection models, we provide both qualitative and quantitative analysis in depth.

We first divide the objects into three groups based on their distance to the ego-car: within 30 meters, from 30 to 50 meters, and beyond 50 meters. Figure~\ref{fig:breakdown} shows the \textit{relative} gain by the multi-modal fusion of each group. 
In a nutshell, \ourproposedmethod{} can uniformly improve the accuracy in every single distance range. In particular,
it can achieve much more accuracy gains for long-range objects (by 6.6\% for LEVEL\_2 objects $>50m$) than short-range objects (by 1.5\% for LEVEL\_2 objects $<30m$), possibly because long-range objects are often covered by very sparse \lidar{} points, but the high-resolution \camera{} signals fill the information gap to a large extent.

Then, we visualize the attention map of \LearnableAlign{} in Figure~\ref{fig:vis_attention}. We observe that model tends to focus on the regions with strong discriminative power such as the head of the pedestrian, and on the object extremities such as the back of the pedestrian. Base on these observations, we conclude that the high-resolution \camera{} signals could help with recognizing, and predicting the object boundaries.

\subsection{Impact of \InverseAug{} and \LearnableAlign{}}

\begin{table}[t]
\begin{center}
\small
\begin{tabular}{l|cc|cccc}
\hline
\multirow{2}{*}{Modal}   & \multirow{2}{*}{IA} & \multirow{2}{*}{LA} & \multicolumn{2}{c}{LEVEL 1} & \multicolumn{2}{c}{LEVEL 2} \\
  &  & & AP  & APH & AP & \textbf{APH} \\\hline
\Lidar{} & N/A & N/A & 78.5 & 70.2 & 70.7 & 63.0 \\
\Lidar{} + \Camera{} & \xmark & \cmark & 78.4 & 70.5 & 70.9 & 63.5  \\
\Lidar{} + \Camera{} & \cmark & \xmark & 80.4 & 74.0 & 72.3 & 66.4  \\
\Lidar{} + \Camera{} & \cmark & \cmark & \textbf{81.2} & \textbf{75.0} & \textbf{72.8} & \textbf{67.0} \\
\hline
\end{tabular}
\end{center}
\vskip -0.15in
\caption{Ablation studies on \InverseAug{} (IA) and \LearnableAlign{} (LA). Both techniques contribute to the performance gain, while \InverseAug{} carries more weights.}
\label{tab:ablation_inverseaug}
\end{table}

In this section, we show the effectiveness of both proposed components, \InverseAug{} and \LearnableAlign{}. According to Table~\ref{tab:ablation_inverseaug}, we observe that both components can improve the the performance over the single-modal baseline. In particular, the gain by \InverseAug{} is more prominent. For example, without \InverseAug{}, the performance of LEVEL\_2 detection drops drastically from 67.0 APH to 63.5 APH, which is already very close to the performance of the \lidar{} only model, 63.0 APH. On the other hand, though milder, the improvement by \LearnableAlign{} is not negligible. For example, \LearnableAlign{} improves the final performance of LEVEL\_2 objects from 66.4 APH to 67.0 APH. 
The ablation study indicates both components are so critical that we should remove neither of them.

\subsection{\ourproposedmethod{} is an effective fusion strategy}

\begin{table}[tb]
\begin{center}
\small
\begin{tabular}{l|c|cccc}
\hline
\multirow{2}{*}{Model} & Latency & \multicolumn{2}{c}{LEVEL 1} & \multicolumn{2}{c}{LEVEL 2} \\
  & (sec) & AP  & APH & AP & \textbf{APH} \\\hline
Single-Modal & 0.16 & 78.5 & 70.2 & 70.7 & 63.0 \\
InputFusion & 0.28 & 79.4 & 74.1 & 72.1 & 66.6 \\
LateFusion & 0.35 & 79.8 & 73.7 & 72.3 & 66.5 \\
\ourproposedmethod{} (ours) & 0.32 & \textbf{81.2} & \textbf{75.0} & \textbf{72.8} & \textbf{67.0} \\
\hline
\end{tabular}
\end{center}
\vskip -0.15in
\caption{Comparison with other fusion strategies. InputFusion comes from PointPainting~\cite{vora2020pointpainting} and PointAugmenting~\cite{wang2021pointaugmenting}. LateFusion comes from PointAugmenting~\cite{wang2021pointaugmenting}. All latency are measured on a V100 GPU with the same Lingvo~\cite{shen2019lingvo} 3D object detection implementation, same 3D detection backbone, and same \camera{} feature extractor. \ourproposedmethod{} achieves the best performance on all evaluation metrics, while the latency is comparable to other fusion methods.}
\label{tab:ablation_fusion}
\end{table}

In this section, we compare \ourproposedmethod{} with other fusion strategies. Specifically, the methods we consider are (1) InputFusion, that fuses the camera features with the \lidar{} points at the input stage~\cite{vora2020pointpainting,wang2021pointaugmenting}, (2) LateFusion, where 
the \lidar{} points and \camera{} features
go through voxelizer separately, followed by a concatenation~\cite{wang2021pointaugmenting}, and (3) our proposed \ourproposedmethod{}.

The results are shown in Table~\ref{tab:ablation_fusion}. We observe that \ourproposedmethod{} is notably better than other fusion strategies. For example, \ourproposedmethod{} improves 0.5 LEVEL\_2 APH (from 66.5 to 67.0) upon LateFusion. Note that in our experiments, InputFusion is on par with LateFusion, 
but in~\cite{wang2021pointaugmenting}, LateFusion is better as it addresses the modality gap issue between \lidar{} and \camera{}. 
We hypothesis that 
in our setting, the modality gap issue is already taken care of by the end-to-end training that it will no longer occur regardless when the fusion is conducted.

\subsection{\ourproposedmethod{} is more robust}
Robustness is an important metric to deploy models on autonomous driving cars ~\cite{michaelis2019benchmarking}.
In this subsection, we examine the model robustness against corrupted input~\cite{hendrycks2018benchmarking} and Out-Of-Distribution (OOD) data~\cite{vyas2018out}.

\begin{table}[tb]
\setlength\tabcolsep{4pt}
\small
\begin{center}
\begin{tabular}{l|c|cccc}
\hline
\multirow{2}{*}{Corruptions} & \multirow{2}{*}{Modal} & \multicolumn{2}{c}{LEVEL 1} & \multicolumn{2}{c}{LEVEL 2} \\
  & & AP & APH & AP & \textbf{APH} \\\hline
No Corruption & L & 78.5 & 70.2 & 70.7 & 63.0~~~~~~~~~~~~ \\
Laser Noise & L & 69.8 & 59.8 & 61.3 & 52.3 {\color{orange}(\textbf{-10.7})} \\ \hline 
No Corruption & L + C & 81.2 & 75.0 & 72.8 & 67.0~~~~~~~~~~~~ \\
Laser Noise & L + C & 81.1 & 74.8 & 72.6 & 66.8 {\color{orange}(\textbf{-0.2})}~~ \\ 
Pixel Noise & L + C & 80.9 & 74.6 & 72.3 & 66.5 {\color{Highlight}(\textbf{-0.5})}~~ \\
Pixel + Laser Noise & L + C & 80.9 & 74.7 & 72.4 & 66.6 {\color{cyan}(\textbf{-0.4})}~~ \\
\hline
\end{tabular}
\end{center}
\vskip -0.15in
\caption{Model robustness against input corruptions. Given the same well-trained single-modal (\textbf{L}idar) and multi-modal (\textbf{L}idar + \textbf{C}amera) models, we evaluate on the original Waymo validation set (NoCorruption), and manually erode the samples from validation set by Laser and Pixel Noise. For Laser Noise, we add perturbation to the reflection values of all laser points. For Pixel Noise, we add perturbation to the \camera{} images. Note that Pixel Noise is only applicable to multi-modal models which use \camera{} images as input. The perturbations are sampled from a uniform distribution with at most 2.5\% of the original value for both Laser and Pixel Noise corruptions.  We observe \ourproposedmethod{} is more robust to these corruptions compared to the single-modal version. \texttt{L} denotes \lidar{}-only; \texttt{L+C} denotes \lidar{} + \camera{}.}
\label{tab:bad_lidar_robustness}
\end{table}

\paragraph{Robustness against input corruptions.} We first test the model robustness on validation set against common corruptions for both modality, Laser Noise (randomly adding noise to \lidar{} reflections) and Pixel Noise (randomly adding noise to \camera{} pixels). For single-modal models, only Laser Noise is applicable while both Laser Noise and Pixel Noise are applicable for multi-modal models. As shown in Table~\ref{tab:bad_lidar_robustness}, with the presence of corruptions, the multi-modal models in general are much more robust than their single-modal counterpart. 
Notably, the Laser / Pixel Noise corruption only can hardly drag down the performance of our multi-modal method (with only 0.2 / 0.5 L2 APH drop). Even when applying both Laser and Pixel Noise corruptions, the performance drop is still marginal (0.4 L2 APH drop). Meanwhile, the single-modal model drops more than 10 APH by simply applying the Laser Noise corruption.

\begin{table}[tb]
\setlength\tabcolsep{5.7pt}
\small
\begin{center}
\begin{tabular}{l|c|cccc}
\hline
\multirow{2}{*}{Validation Set} & \multirow{2}{*}{Modal} & \multicolumn{2}{c}{LEVEL 1} & \multicolumn{2}{c}{LEVEL 2} \\
  & & AP & APH & AP & \textbf{APH} \\\hline
Default & L & 78.5 & 70.2 & 70.7 & 63.0~~~~~~~~~~~ \\
Default & L + C & 81.2 & 75.0 & 72.8 & 67.0\improves{+4.0} \\\hline
Kirkland & L & 43.8 & 38.8 & 31.2 & 27.6~~~~~~~~~~~ \\
Kirkland & L + C & 52.0 & 47.4 & 38.0 & 34.6\Improves{+8.0} \\\hline
\end{tabular}
\end{center}
\vskip -0.1in
\caption{Model robustness against out-of-distribution data. We evaluate both single-modal (\textbf{L}idar) and multi-modal (\textbf{L}idar + \textbf{C}amera) models on the in-distribution validation set (Default) and out-of-distribution validation set (Kirkland). \ourproposedmethod{} achieves larger improvement on out-of-distribution validation set. \texttt{L} denotes \lidar{}-only; \texttt{L+C} denotes \lidar{} + \camera{}.}
\label{tab:ood_robustness}
\end{table}

\paragraph{Robustness against OOD data.} To test our method's robustness against OOD data, we train our model on the data from cities of Mountain View, San Francisco and Phoenix and evaluate it on Kirkland. The results are summarized in Table~\ref{tab:ood_robustness}. We observe the multi-modal model shows more improvement on OOD data. For example, \ourproposedmethod{} improves 8.0 LEVEL\_2 APH on out-of-distribution data while only improves 4.0 LEVEL 2 APH on in-distribution data.

\section{Conclusion}
This paper studies how to effectively fuse \lidar{} and \camera{} data for multi-modal 3D object detection. Our study shows deep feature fusion in late stage can be more effective when they are aligned well, but aligning two deep features from different modality are challenging. To address this challenge, we propose two techniques, \InverseAug{} and \LearnableAlign{}, to get effective alignment among multi-modal features. Based on these techniques, we develop a family of simple, generic, yet effective multi-modal 3D detectors, named \ourproposedmethod{}s, which achieves state-of-the-art performance on the Waymo Open Dataset.

\paragraph{Acknowledgement.}  
We would like to thank Zhaoqi Leng and Ekin Dogus Cubuk for data augmentation discussion, and Pei Sun for model ensemble discussion. Yingwei Li would like to thank Longlong Jing for the detailed tutorial on 3D object detection task, and Zhiwen Wang for suggestions on figures.

\appendix

\section{Appendix}
\subsection{Impact of Alignment Quality} 
In this section, we provide more detailed experimental settings and more results of our preliminary experiments in Section 3.2 in the main paper.

\paragraph{Experimental Settings.} We use the 3D-MAN++ Pedestrian model mentioned in Section 4.1 and Section~\ref{sec:app_imp_detail}. To examine the alignment quality, \InverseAug{} and all data augmentations are removed. Then, we apply different magnitude of RandomRotation~\cite{zhou2018voxelnet} to both single-modal and multi-modal models. Finally, for the same perturbation magnitude, we compute the performance gap for the best validation results from the single-modal and multi-modal models.

\begin{figure*}[tb]
\centering
\includegraphics[width=0.95\linewidth]{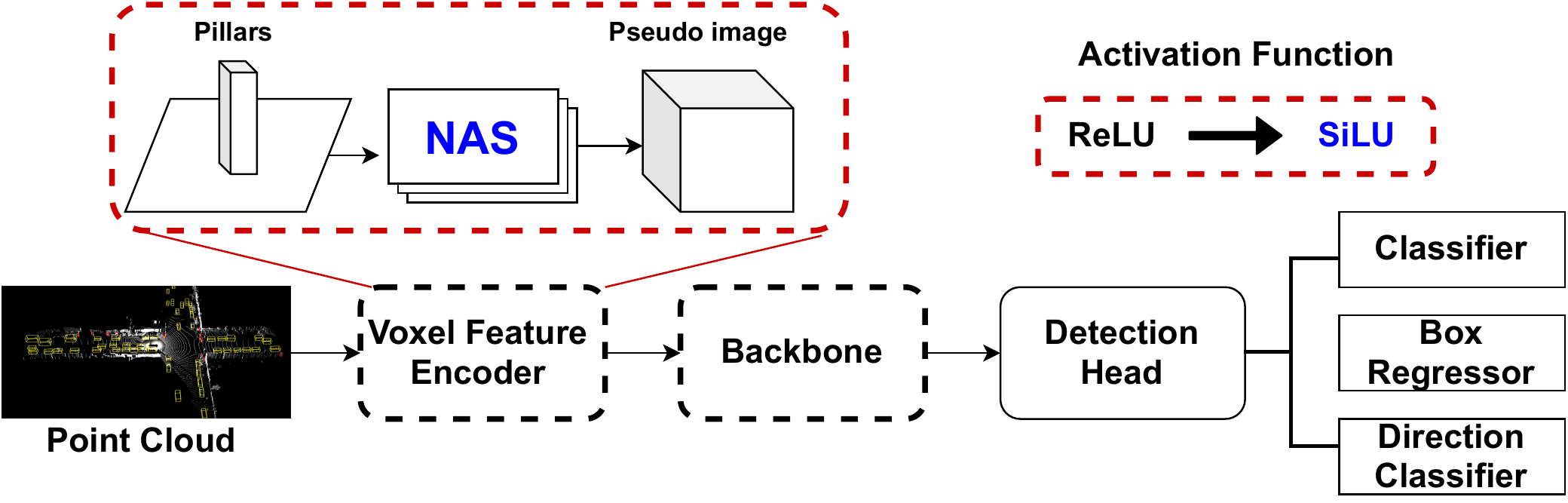}
\caption{The overview of PointPillars framework and its improved implementation (marked in red dashed boxes). Our improved implementation replace the original Voxel Feature Encoder from one fully-connected layer to a multilayer perceptron, whose hyperparameter (such as the number of layers, and the number of filters) are automatically discovered by Neural Architecture Search~\cite{zoph2016neural}. In addition, we change the non-linear activation function from ReLU~\cite{hahnloser2000digital,nair2010rectified} to SILU~\cite{elfwing2018sigmoid,ramachandran2017searching}.} 
\label{fig:improved_implementation}
\end{figure*}

\begin{figure*}[tb]
\begin{center}
  \includegraphics[width=0.9\linewidth]{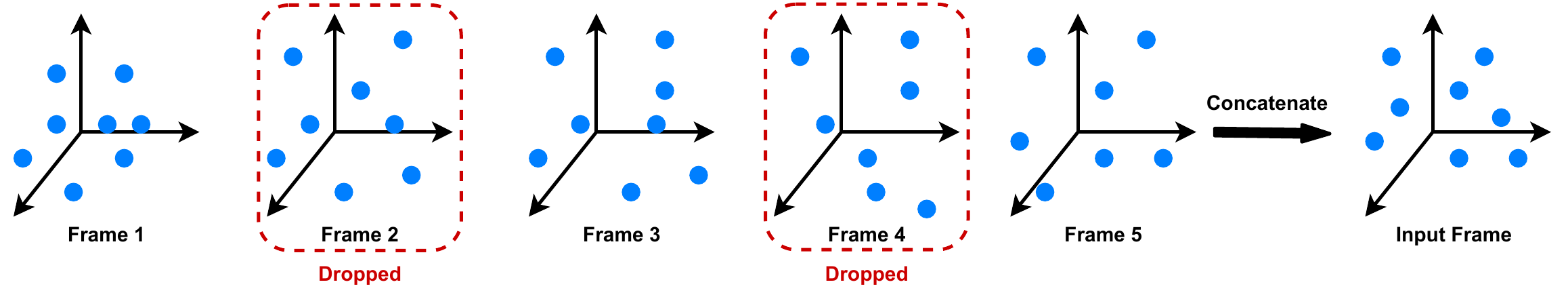}
\end{center}
  \caption{The DropFrame process during training of a 5-frame model. In this example, Frame 2 and 4 are dropped out before collapsing. The frames selected to be dropped are randomly selected in each training step. During inference, all frames are used as input without any frames dropped.}
\label{fig:dropframe}
\end{figure*}

\paragraph{Additional Results.} Besides testing with RandomRotation~\cite{zhou2018voxelnet}, we also test with RandomFlip~\cite{zhou2018voxelnet}, another commonly used data augmentation strategy for 3D point cloud object detection models. Specifically, RandomFlip flip the 3D scene along the Y axis with a given probability $p$. Here, we set the probability as 0\%, 50\%, and 100\%, respectively, and the results are shown in Table~\ref{tab:alignment_flip}. The observation is similar: when applying large magnitude data augmentation, the benefit from multi-modal fusion diminishes. For example, when applying zero-probability RandomFlip (\ie, not applying this data augmentation), the improvement is the most significant (+2.3 AP); when flip probability is 100\% (\ie, flip the 3D scene every time), the improvement is almost zero (+0.03 AP).

\subsection{Implementation Details of 3D Detectors} \label{sec:app_imp_detail}
In the main paper, we mainly focus on providing more details about \ourproposedmethod{} due to the space limitation. In this section, we will also illustrate other important implementation details to build the strong 3D object detection models.

\paragraph{Point cloud 3D object detection methods.} We reimplement three popular point cloud 3D object detection methods, PointPillars~\cite{lang2019pointpillars}, CenterPoint~\cite{yin2021center}, and 3D-MAN~\cite{yang20213d}. As mentioned in Section 2, PointPillars voxelize the point cloud by pillars, a single tall elongated voxel per map location, to construct bird eye view pseudo image; finally, the pseudo image is fed to an anchor-based object detection pipeline. A high-level model pipeline is shown in Figure~\ref{fig:improved_implementation}. CenterPoint is also a pillar-based method, but using anchor-free detection head instead. Note that we only implemented the PointPillars-based single-stage version of CenterPoint. 3D-MAN is similar to CenterPoint, and the main difference is when computing the loss, 3D-MAN uses a Hungarian algorithm to associate the prediction and the ground-truth (See Section 3.1 of Yang~\etal~\cite{yang20213d} for more details). 

\begin{table}[tbh]
    \centering
    \begin{tabular}{l|c|c|c}
    \hline
         Flip Probability & 0\% & 50\% & 100\%  \\
         \hline
         Single-Modal & 72.6 & 76.7 & 71.8 \\
         Multi-Modal & 74.9 & 76.8 & 71.9  \\
         Improvement & +2.3 & +0.10 & +0.03 \\
         \hline
    \end{tabular}
    \caption{Performance gain by multi-modal fusion diminishes as the magnitude of RandomFlip~\cite{zhou2018voxelnet} goes up, indicating the importance of accurate alignment. \InverseAug{} is not used here. On the Waymo Open Dataset pedestrian detection task, the LEVEL 1 AP improvements from single-modal to multi-modal are reported.}
    \label{tab:alignment_flip}
\end{table}

\paragraph{Our improved implementations.} We also introduce two simple but effective findings that significantly improve the point cloud 3D object detection baselines. We take the PointPillars framework as an example to introduce them, but these techniques can be naturally applied to other point cloud 3D object detection frameworks, such as CenterPoint and 3D-MAN. As shown in Figure~\ref{fig:improved_implementation}, we build upon the PointPillars model and indicate our modifications the red dotted line boxes. The NAS block depicts the voxel feature encoder found using architecture search. We also replace the ReLU~\cite{hahnloser2000digital,nair2010rectified} activation function in the original frameworks with SILU~\cite{elfwing2018sigmoid,ramachandran2017searching}. Our improved models (named as PointPillars++, CenterPoint++, and 3D-MAN++) shows better performance that its baseline method as shown in Table 4 in the main paper. For example, for 3D-MAN, after applying these two techniques, the LEVEL\_2 APH is improved from 52.2 to 63.0. This improvement is significant, and is consistently observed from other metrics and from other baselines.

\paragraph{Other training details.} We use both LEVEL\_1 and LEVEL\_2 difficulty data for training. Since the LEVEL\_2 data is difficult for model to predict, we use the uncertainty loss~\cite{meyer2020learning} during training to tolerate the models to detect low-confident objects with low accuracy. 

\paragraph{Details of the submitted models.}
We apply \ourproposedmethod{} to CenterPoint to prepare our models for submission. We enlarge the Max Rotation for the RandomRotation augmentation to 180$^{\circ}$ (120$^{\circ}$ for Pedestrian model) since we see its benefit according to Table~\ref{tab:alignment}. We also enlarge the pseudo-image feature resolution from 512$\times$512 to 704$\times$704.
We combine the information from previous frames by simply concatenating the point-clouds across the last $N$ frames together. As shown in Figure~\ref{fig:dropframe}, to prevent the over-fitting issue under the multi-frame setting, we propose DropFrame, that randomly drop the point cloud from previous frames. Our very best model concatenates 5 frames, and with dropframe probabilities 0.5 during training. Besides, we  also apply model ensemble and Test-Time Augmentation (TTA) by weighted box fusion (WBF)~\cite{hu2021afdetv2}. For TTA, we use yaw rotation, and global scaling. Specificly, we use [0$^{\circ}$, $\pm$22.5$^{\circ}$, $\pm$45$^{\circ}$, $\pm$135$^{\circ}$, $\pm$157.5$^{\circ}$, $\pm$180$^{\circ}$] for yaw rotation, and [0.95, 1, 1.05] for global scaling. For model ensemble, we obtain 5 different type of models with different pseudo-image feature resolution and different input modality,~\ie, single-modality 512 / 704 / 1024 resolution, and multi-modality 512 / 704 resolution. For each type of model, we train 5 times with different random seed. Then, we rank all 25 models with the performance on validation set and ensemble top-k models, where k is the optimal value to get the best results on the validation set.

\subsection{Comparison with Larger Single-Modal Models}

\begin{figure}[tbh]
\centering
\includegraphics[width=0.95\linewidth]{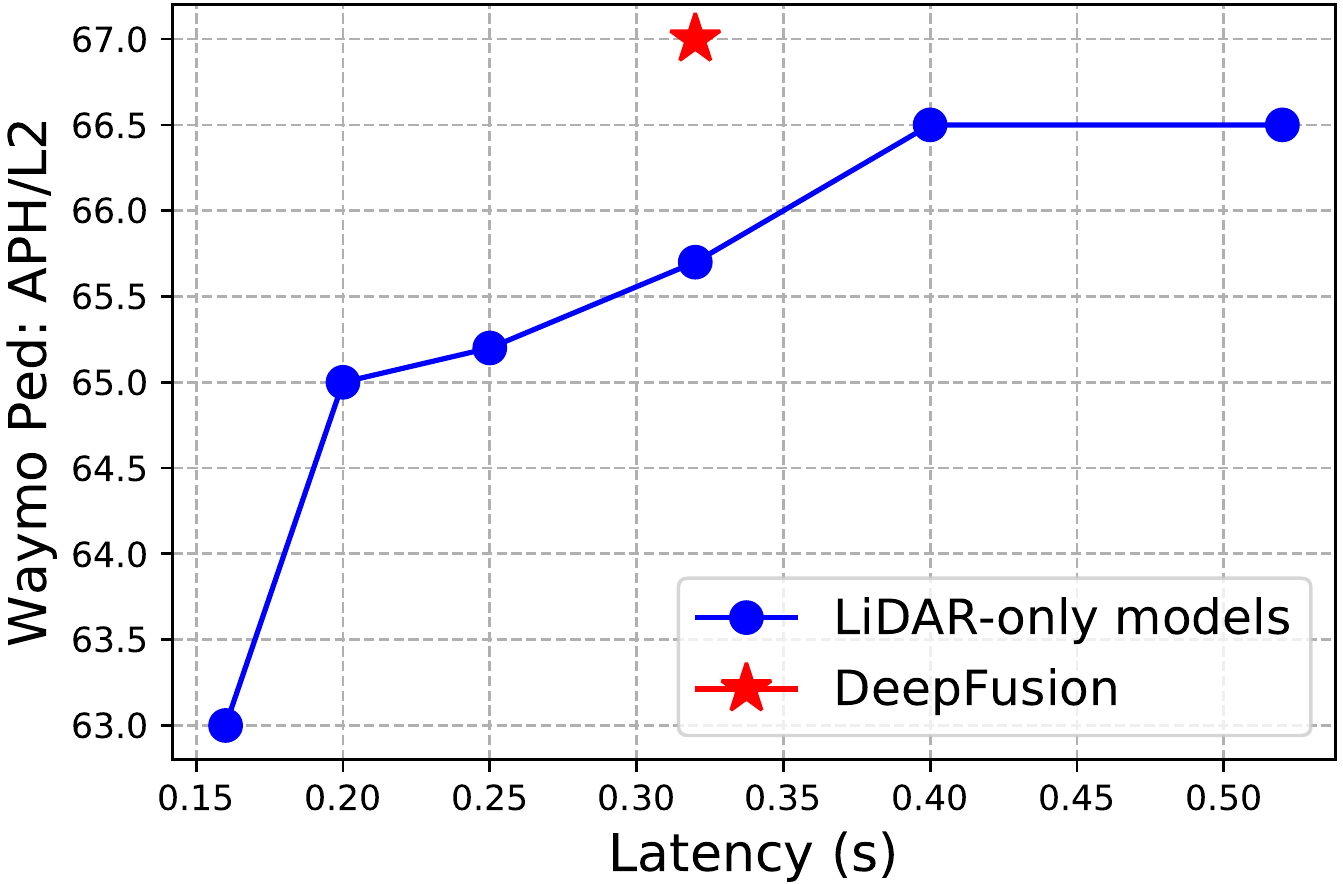}
\vskip -0.1in
\caption{Model latency \textit{vs.} detection performance. 
\ourproposedmethod{} significantly outperforms single-modal models under all latencies.
}
\label{fig:scaleup}
\vskip -0.1in
\end{figure}

The goal of this section is to compare the Single-Modal baseline with \ourproposedmethod{} under the same computational budget. To achieve this, we first scale up the single-modality model.
Since we have sufficiently scaled up the  voxel feature encoder and backbone when building the baseline models, enlarging the resolution of the pseudo image is probably the most effective way for further scaling the single-modal model to match the latency with multi-modal model, and thus we adopt such a strategy here.  
Specifically, we train the models under resolutions ranging from 512 up to 960, and test the performance of each setting. 
Figure~\ref{fig:scaleup} clearly shows that \ourproposedmethod{} achieves 67.0 L2 APH with 0.32s latency while the single-modal model can only achieve 65.7 L2 APH with the same latency budget. 
Further scaling up the single-modal model brings marginal gain to the performance, which is capped at 66.5 L2 APH and still worse than \ourproposedmethod{}.

\paragraph{Limitations:} This paper focuses on fusing \lidar{} and \camera{} information. However, our proposed method could also be potentially extended to other modalities, such as range image, radar and high-definition map. Besides, we simply adopt voxel-based methods such as PointPillars~\cite{lang2019pointpillars}, but it is possible to further improve the performance by adopting strong baselines~\cite{sun2021rsn}. 

\paragraph{License of used assets:} Waymo Open Dataset~\cite{sun2020scalability}: Waymo Dataset License Agreement for Non-Commercial Use (August 2019). \footnote{\url{https://waymo.com/intl/en_us/dataset-download-terms/}}

{\small
\bibliographystyle{ieee_fullname}
\bibliography{egbib}
}

\end{document}